\documentclass[letterpaper, 10 pt, conference]{ieeeconf}
\IEEEoverridecommandlockouts

\usepackage{cite}
\usepackage{amsmath,amssymb,amsfonts}
\usepackage{algorithmic}
\usepackage{graphicx}
\usepackage{textcomp}
\usepackage{xcolor}
\usepackage{siunitx}
\usepackage{float}
\usepackage{url}
\usepackage{dblfloatfix}
\usepackage{orcidlink}

\definecolor{tumblue}{RGB}{0,101,189}
\definecolor{darkbluetum}{RGB}{0,82,147}
\definecolor{darkerbluetum}{RGB}{0,51,89}
\definecolor{orangetum}{RGB}{227,114,34}
\definecolor{greentum}{RGB}{162,173,0}
\definecolor{lightbluetum}{RGB}{152,198,234}
\definecolor{pastelbluetum}{RGB}{100,160,200}

\usepackage{textcomp}

\begin{document}

\title{\LARGE \bf SPARK: Low Latency Single-Camera 3D Pose Estimation for Autonomous Racing using Keypoints}

\author{Dominic Ebner and Markus Lienkamp\thanks{All authors are with the Technical University of Munich, Germany; School of Engineering \& Design, Department of Mobility Systems Engineering, Institute of Automotive Technology and Munich Institute of Robotics and Machine Intelligence (MIRMI) \newline{}
    Corresponding author: \href{mailto:dominic.ebner@tum.de}{dominic.ebner@tum.de}}}
\maketitle
\thispagestyle{empty}
\pagestyle{empty}

\begin{abstract}
  In autonomous racing, fast detection of other participants' movements is required to plan safe, collision-free trajectories with non-cooperative opponents. LiDAR detection is inherently slower and harder to deploy on edge devices than vision methods, causing delayed detections that limit object tracking performance during high-dynamic maneuvering.
  Utilizing monocular 3D detection enables an easy-to-deploy, low-latency detection of other participants on the racetrack.
  We present SPARK, a single-camera pose-estimation algorithm for autonomous racing using keypoint detection.
  It achieves long-range detection with high accuracy, exceeding the performance of state-of-the-art monocular camera detection algorithms while maintaining lower latency.
  By employing well-optimized YOLO models and leveraging the fixed geometry in the autonomous racing domain, the algorithm also exhibits low latency and resource usage.
  We evaluate the performance of our approach on real-world autonomous racing data and compare it to state-of-the-art LiDAR and camera detection algorithms.
  The source code is available at: \url{https://github.com/TUMFTM/SPARK-camera-det}
\end{abstract}

\section{Introduction}

In autonomous racing, as in real racing, split-second decisions matter when facing non-cooperative opponents in high-dynamic scenarios \cite{iac_indy_nodate,aspire_abu_nodate,wischnewski_indy_2022,hoffmann_head--head_2026}.
Sudden changes in speed and direction are exaggerated due to the vehicle's capabilities.
Therefore, perception algorithms need extremely low latencies and long range to enable the following software stack to make the best possible decisions for both performance and safety.
Combining all available sensor modalities is advantageous, but limited by the constrained compute capability of the autonomous vehicle.

The biggest advantage of lower latencies is the reduced delay compensation and prediction required to track opponents in real time.
Extended-Kalman-Filter-based object tracking \cite{raji_erautopilot_2024,karle_multi-modal_2023} benefits from lower latencies, especially in autonomous racing, as the vehicle can achieve up to 30\unit{\meter\per\second\squared} of acceleration during interaction.
High relative velocities and yaw-rates can appear almost instantaneously, and tracking this movement is improved by accurate, low-latency, high-frequency detections.

Current LiDAR sensors capture data at 10-30\unit{\hertz} \cite{ouster_overview_nodate,hesai_pandar128_nodate,luminar_luminars_nodate,seyond_falcon_nodate}, which can cause significant delays in object detection using these pipelines.
CenterPoint \cite{yin_center-based_2021} reports detection speeds of 11\unit{\hertz} on the Waymo dataset and 16\unit{\hertz} on the nuScenes \cite{caesar_nuscenes_2020} dataset, which is not enough to sequentially handle every point cloud produced by a 20\unit{\hertz} LiDAR, but also introduces latencies of 60-90\unit{\milli\second} to detected objects.
VoxelNeXt \cite{chen_voxelnext_2023} addresses this issue with sparse convolutions, but also does not report latencies below the necessary 50\unit{\milli\second} for real-time detection using 20 \unit{\hertz} LiDARs.
While latencies are rarely reported on similar hardware, LiDAR detection algorithms rarely achieve latencies below 10 \unit{\milli\second}.

\begin{figure}
    \centering
    \includegraphics[width=1.0\linewidth]{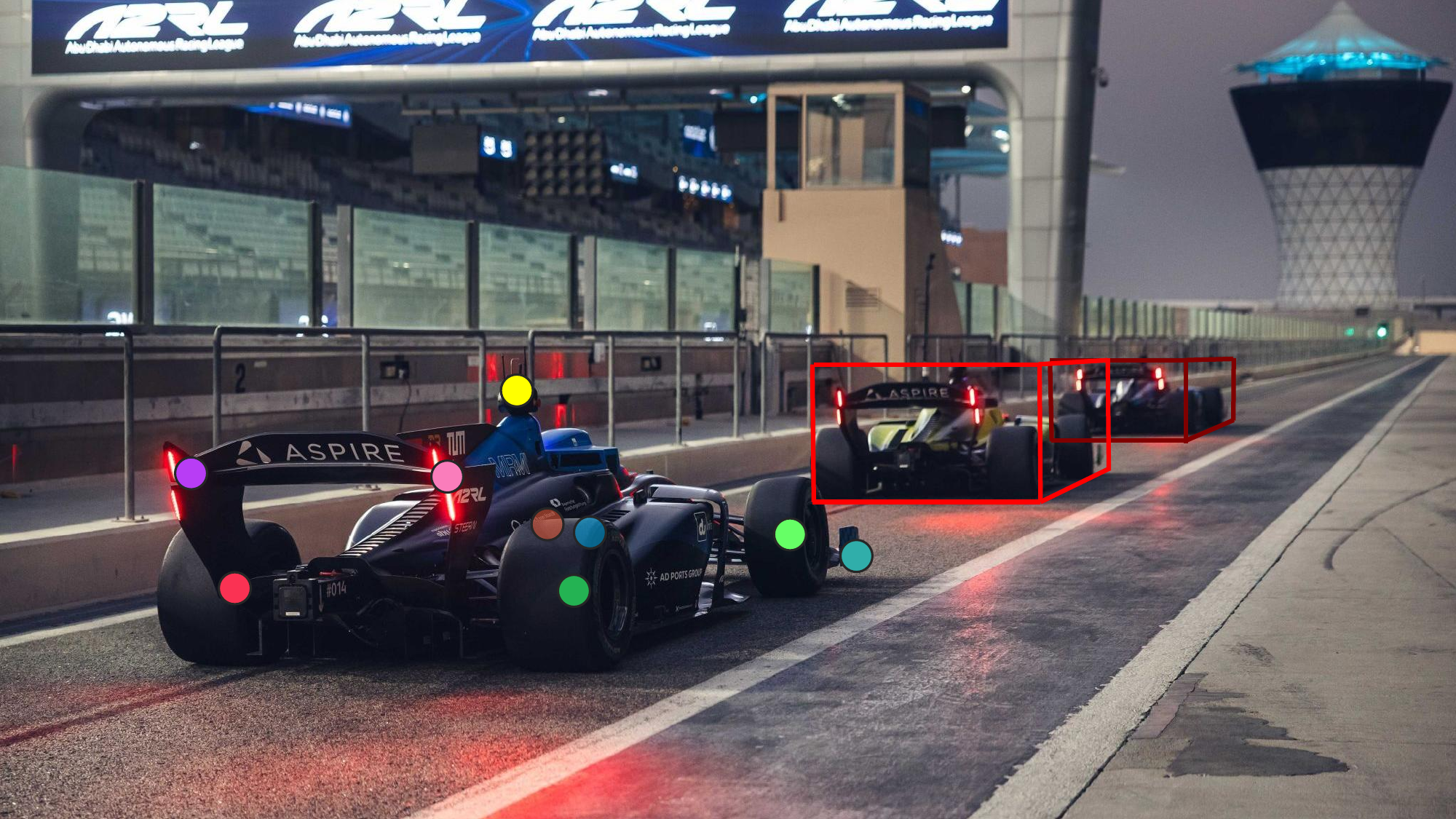}
    \caption{Multi-Vehicle autonomous racing on the Yas Marina Circuit during the Abu Dhabi Autonomous Racing League (A2RL) \cite{aspire_abu_nodate}. The vehicles reached up to 260\unit{\kilo\meter\per\hour} during interaction. Photo credit: Aspire}
    \label{fig:hero-shot}
\end{figure}

Compared to LiDAR, lightweight vision networks are a natural choice for low-latency object detection \cite{tian_yolov12_2025,ultralytics_ultralytics_nodate}.
While they are fast \cite{liao_kitti-360_2023,kitti-360_kitti-360_nodate}, they struggle with accurate depth regression.
To recover depth information, complex methods have been introduced that combine 2D backbones with 3D detection heads \cite{huang_monodtr_2022,zhang_monodetr_2023,yan_monocd_2024,liu_smoke_2020,simonelli_disentangling_2019}.

However, unlike road applications, in autonomous racing, there is only a single target class with known dimensions.
Leveraging knowledge of this constrained domain can help accelerate and improve object detection.
Our main contributions are:
\begin{enumerate}
    \item We introduce SPARK, an open-source, lightweight, low-latency 3D object-detection algorithm for objects with known, fixed dimensions.
    \item We present and open-source a method to generate highly accurate 2D keypoint annotations from 3D LiDAR ground truth annotations.
    \item We benchmark the algorithm's detection performance against state-of-the-art monocular 3D detection methods and LiDAR object detection methods on high-speed autonomous racing data.
\end{enumerate}
 
\section{Related Work}

\subsection{Latency of perception algorithms}

Object detection evaluation in autonomous driving primarily focuses on Mean Average Precision (mAP) on benchmark datasets.
Detection and tracking benchmarks like nuScenes do not evaluate detection latency at all when producing metrics \cite{caesar_nuscenes_2020}.
Current voxel- or pillar-based algorithms report inconsistent runtimes across different hardware, while typically staying below a throughput of 30\unit{\hertz}.
Detection latency is also dependent on the voxel grid size, which increases quadratically when using pillars or fixed-height voxel grids with dense convolutions\cite{chen_voxelnext_2023,fan_fully_2022}.
Additionally, algorithm runtime is rarely evaluated using identical hardware, therefore prohibiting direct comparisons between claims.

PointPillars has been the target of runtime improvements due to its simple, dense backbone, enabling deployment and optimization with ONNX \cite{onnx_onnxonnx_2026} and TensorRT \cite{nvidia_nvidiatensorrt_2026,nvidia_nvidia-ai-iotcuda-pointpillars_2026}.
Implementations using GPU-accelerated voxelization, optimized inference, and post-processing can reach 100\unit{\hertz} on small voxel grids.
Sparse backbones are difficult to deploy with ONNX and TensorRT, with limited support in specific frameworks \cite{tier4_tier4awml_2026}.
Of the current state-of-the-art detection models, only CenterPoint is commonly deployed for accelerated inference \cite{openmmlab_mmdet3d_2020}, however it requires splitting the model into multiple sub-models and custom implementations for operations not supported by ONNX.

2D object detection usually exhibits significantly lower latency\cite{ultralytics_ultralytics_nodate}; however, lifting the detection from 2D into 3D remains challenging.
3D object detection shows promising latency results on powerful hardware, reporting latencies $<$5\unit{\milli\second} on the KITTI-360 benchmark using current high-performance compute hardware\cite{kitti-360_kitti-360_nodate}.

\subsection{Depth Estimation}

Due to the limited domain in autonomous racing, full regression of 3D bounding boxes from 2D features is not strictly necessary. Given known dimensions, a distance can be estimated from the 2D bounding box of a simple detection model. Using a known height $H$, a focal length $f$ of a pinhole camera model, and the pixel height of the object $h$, a distance $Z$ can be estimated.
This estimation, however, depends on the relative pose of the opponent vehicle, which cannot be determined from simple 2D bounding boxes. Detection using this method was previously evaluated and found to be too inaccurate for use in autonomous racing \cite{wischnewski_indy_2022}.
While a 3D position can be estimated, a 6D pose cannot be generated using this process, as the object's orientation is not considered.
This simple estimation is also used in MonoDis \cite{simonelli_disentangling_2019} to generate a depth estimation after regressing the height $H$.

\subsection{Perspective-n-Point Problem}

Given a projection matrix $K$, a set of image coordinates $p_c = (u_i,v_i, 1)$, and world coordinates $p_w = (x_i,y_i,z_i)$, a 6 Degrees-of-Freedom (DoF) pose can be recovered an image \cite{marchand_pose_2016}.
The task of the Perspective n' Point (PnP) algorithm is to determine a rotation $R$ and translation $t$ so that:

\begin{align}
    p_c = K \cdot ( R \cdot p_w + t)
\end{align}

While this problem definition assumes an undistorted image, the distortion coefficients of the camera can also be used in the open-source OpenCV implementation to solve the PnP problem on distorted images \cite{opencv_perspective-n-point_nodate}.

\subsubsection{P3P \cite{zhang_general_2005}}

With exactly three image and world coordinates, the PnP problem can be solved using \textit{P3P}, yielding 4 solutions. This solution is not viable for determining a vehicle's orientation without making specific assumptions about its most likely pose.

\subsubsection{EPnP \cite{lepetit_epnp_2009} \& SQPnP \cite{terzakis_consistently_2020}}

Efficient PnP (EPnP) solves the PnP problem with at least four image and world coordinates. It approximates a single $O(n)$- time solution by solving a linearized system with Gauss-Newton refinement.
SQPnP finds a globally optimal solution by solving the PnP problem as a quadratic constrained optimization problem. It is slower than EPnP, but promises a more accurate solution.

\subsubsection{RANSAC}

Random-Sampling-Consensus (RANSAC) can be used to improve pose estimation in the PnP problem by iteratively removing outliers. EPnP or SQPnP approaches are used to compute an initial solution, and reprojection errors are used to determine and filter outliers.

\subsection{2D Keypoint Detection}

Pose detection models are traditionally trained to detect human poses, and many of them also exhibit extremely low latency and resource usage \cite{li_rethinking_2019,mmpose_contributors_openmmlab_2020,jiang_rtmpose_2023}.
There are numerous approaches for this domain; however, they often lack flexibility in the input image size and the target keypoints.
YOLO-Pose models \cite{ultralytics_ultralytics_nodate} are flexible with input size and keypoint definitions, while also offering many options for model size to achieve a desirable latency.
A 6 DoF pose estimation approach using PnP has been presented for the estimation of spacecraft poses by first predicting keypoints \cite{zhang_monocular_2024}.

\subsection{Monocular 3D Object Detection}

\begin{figure*}[t]
    \vspace{0.2cm}
    \centering
    \includegraphics[width=\linewidth]{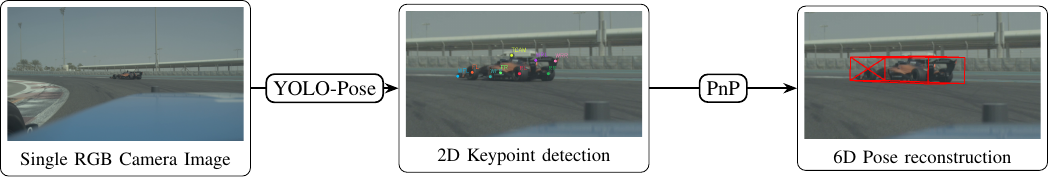}     \label{fig:algo}
    \caption{Our approach for detecting vehicles using only a YOLO-Pose model without a 3D detection head. We first detect 2D vehicle keypoints and then solve the PnP problem using the keypoints' 3D local coordinates. The front face of the red bounding box is marked with an X.}
\end{figure*}

Monocular camera 3D object detection is a challenging task, and there are many diverse approaches to this problem in autonomous driving \cite{liu_smoke_2020,li_rtm3d_2020,huang_monodtr_2022,zhang_monodetr_2023,shi_geometry-based_2021,avidan_petr_2022,yan_monocd_2024,simonelli_disentangling_2019,zou_end--end_2021}.
Full object bounding boxes are predicted from single RGB images, typically on the KITTI \cite{geiger_vision_2013} or nuScenes \cite{caesar_nuscenes_2020} dataset.
MonoDIS \cite{simonelli_disentangling_2019} does not require any additional information beyond 3D annotations for training, making training and deployment on a new dataset easy.
SMOKE \cite{liu_smoke_2020} and MonoCD \cite{yan_monocd_2024} require ground-plane information to estimate object heights, which complicates the application of these algorithms to new data.
Recently, transformers were successfully used to improve detection performance on the KITTI dataset.
MonoDTR \cite{huang_monodtr_2022} uses LiDAR ground-truth depth data to support regression of the 3D detection head.
It relies on precomputed object anchors and center-based object detection.
MonoDETR \cite{zhang_monodetr_2023} uses both image features and a predicted depth map, while improving on MonoDTR by not relying on anchors and utilizing global features for local queries.
RTM3D \cite{li_rtm3d_2020} \& KM3D \cite{li_monocular_2021} both predict object keypoints and center points with geometric constraints to support 3D position regression.
In conclusion, most methods are designed to solve a more complex problem than is required for autonomous racing, where target dimensions are fixed, and there is only one type of object in the scene.
AutoShape \cite{liu_autoshape_2021} uses CAD models to support detection, which shows promise for the specific domain, as a perfectly matching CAD model exists for the racecar.
However, the training pipeline requires a complex preprocessing step, which complicates the application of the algorithm.

Model complexity is rarely reported for these models, and latency is evaluated commonly, but always on non-standard hardware.
Reported latencies for KM3D \cite{shi_geometry-based_2021} promise real-time performance at $\sim$30\unit{\milli\second} on older hardware, which is optimal for autonomous racing; however, detection accuracy is lower than other methods.

\subsection{Datasets}

Most of the discussed monocular detection algorithms are trained on the KITTI dataset \cite{geiger_vision_2013}. It includes LiDAR, calibration data, RGB and depth camera images, and 2D and 3D bounding box annotations.
It consists of 5985 training frames and 1496 validation frames, comprising street-driving scenes in Germany.
Annotations are limited to a distance of 60 meters in front of the ego vehicle, which is insufficient for the autonomous racing domain.

While multiple datasets exist for the autonomous racing domain \cite{nye_betty_2025,kulkarni_racecar_2023}, algorithm performance is never evaluated on these datasets.
In these datasets, annotations are usually automatically generated and cannot be verified for correctness without manual oversight.
In the RACECAR dataset \cite{kulkarni_racecar_2023}, annotations are generated using GNSS positioning, which is prone to errors and requires precise synchronization to produce high-quality relative poses for ground-truth annotations.
While the Betty dataset \cite{nye_betty_2025} claims accurate poses, its data has not yet been released to the public.
Domain shifts also affect this data, limiting its use for training purpose-built algorithms.
For this reason, we needed to create our own dataset with manual annotations for the best possible detection accuracy.

\section{Method}

\subsection{2D Keypoint Detection}

To enable fast 3D object detection from a single RGB image, we combine a deep learning approach with classical pose reconstruction.
We utilize the YOLOv11 \cite{ultralytics_ultralytics_nodate} and YOLOv12 \cite{tian_yolov12_2025} pose detection models for keypoint detection and later recover the pose using the PnP solver.
Combining these approaches leverages the fast inference and efficiency of YOLO-Pose models in 2D while lifting 2D detections into 3D without complex, slow deep-learning methods.

\begin{figure*}[t]
    \centering
    \includegraphics[trim=0cm 3.5cm 0cm 2.5cm,width=0.8\linewidth]{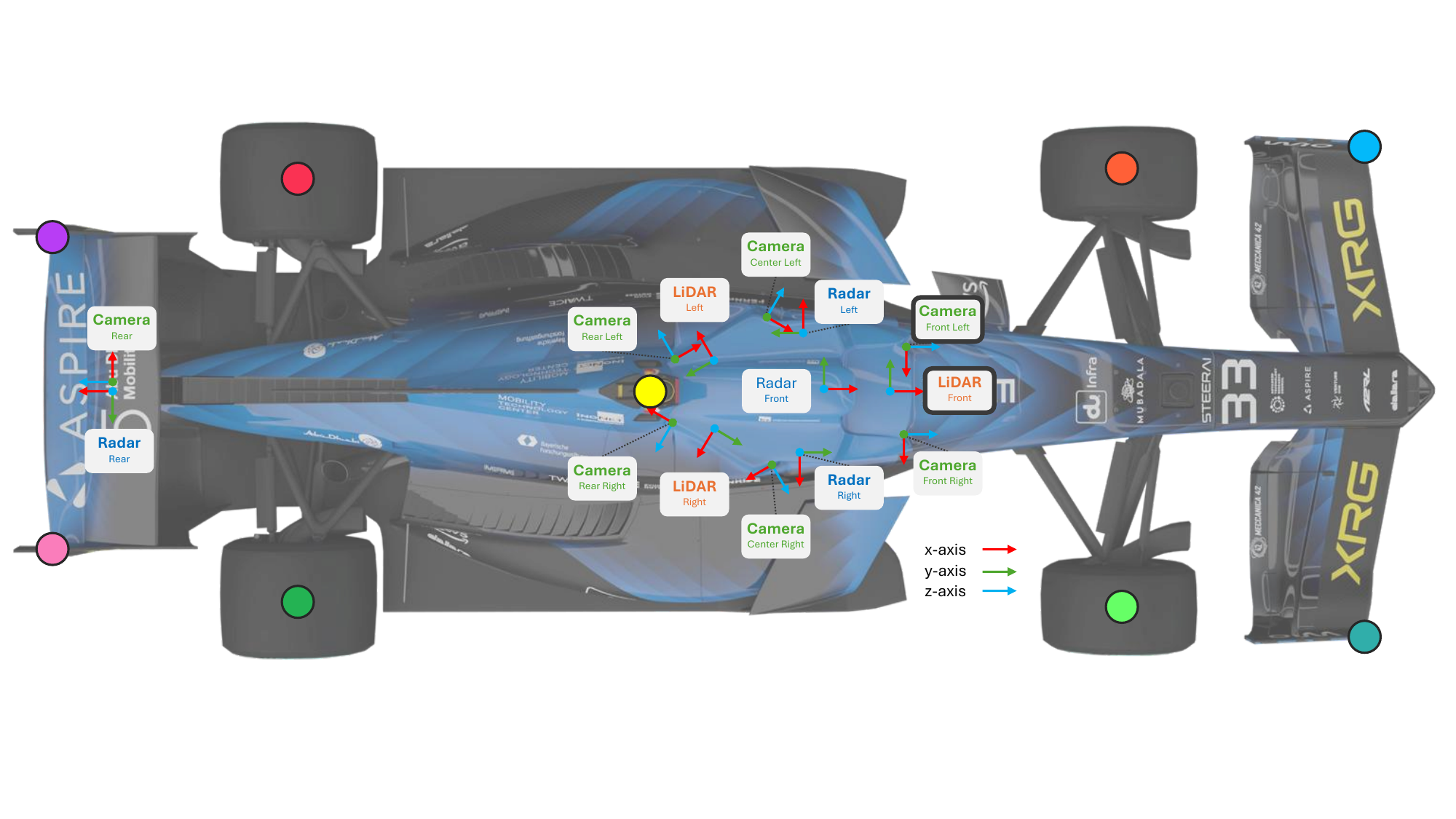}
    \caption{Sensor layout of the EAV24 autonomous racecar. The outlined front-facing camera and LiDAR were used to create the dataset. The selected keypoints for pose detection are shown as colored circles on the tires, wings, and the vehicle's top camera.}
    \label{fig:sensor_layout}
\end{figure*}

To recover a single 6D pose from keypoints, at least four keypoint correspondences need to be passed to the PnP solver \cite{opencv_perspective-n-point_nodate}.
Intuitively, one would choose the four tires as clear indicators of the vehicle's pose.
However, depending on the opponent vehicle's pose, some tires may be occluded, and the requisite four keypoints for EPnP or SQPnP may not be accurately detected.
To prevent this, 9 keypoints were selected so that at least 4 keypoints are always visible in any orientation, enabling accurate and consistent pose detection.
Keypoints may not be visible for multiple reasons: Self-occlusions due to relative heading (left tires are not visible when the other car is viewed from the right), occlusions due to track geometry, and occlusions from other vehicles.
In addition to detecting and individually identifying tires, the vehicle's Top-Camera housing, front-wing edges, and rear-wing edges were annotated as shown in Figure \ref{fig:sensor_layout}.
These keypoints add non-coplanar points to the vehicle's axles and aid detection when parts of the vehicle are occluded.

Because occluded keypoints cannot be manually refined during annotation, they were marked with a third property in addition to image coordinates.
Each annotated keypoint $p$ is defined as $(u,v,vis)$, with $vis$ being 0 for \textit{outside}, 1 for \textit{occluded} and 2 for \textit{visible}.
This property is passed to the YOLO algorithm during training but ignored during inference.
During training of the default YOLO, $vis$ affects the model's loss calculations, but a dummy value is returned during inference.
The YOLO loss calculation was modified to include a Binary Cross-Entropy loss on the $vis$ property.
We encode both \textit{outside} and \textit{occluded} keypoints as 0 and \textit{visible} as 1.
With this change, the model also predicts a usable $vis$ per keypoint, which can later be used to filter points for pose recovery using PnP.

\subsection{Pose Reconstruction}

After 2D keypoint detection, we create image-keypoint-to-object-keypoint correspondences.
Since all keypoints are individually identified, no other information is needed to match image to object keypoints.
The object keypoint positions were measured from a CAD model that matches the vehicle's real measurements exactly.
Pose reconstruction is done using the SQPnP solver as long as there are at least 4 input points.
RANSAC PnP can also be used to iteratively estimate the pose and filter outlier keypoints.
To handle occlusion, we optionally filter keypoints based on predicted $vis$ values.
When parts of the vehicle are occluded, we do not want to rely on the YOLO model's predictions when the visual features for detection are absent from the image.
We define a threshold $t_{vis}$ to filter the set of keypoints $P$ from the input of the PnP solver:

\begin{align}
    P_{filtered} = \{(u_i,v_i)| vis_i \geq t_{vis} \forall (u_i,v_i,vis_i) \in P\}
\end{align}

If occlusions reduce the point set to at least 3 points, no definitive pose can be reconstructed, so the 2D keypoint detection is discarded.
 
\section{Dataset}

\subsection{Sensor Platform}

The sensor layout of the Dallara EAV24 is shown in Figure \ref{fig:sensor_layout}. The vehicle is based on a Dallara SF23 used in the Japanese Super Formula racing series and was converted to an autonomous platform by A2RL.

For this work, only the front-left camera, a Leopard Imaging IMX728-80H, and the front LiDAR, a Seyond Falcon K, were used.
The camera can record images at up to 45\unit{\hertz}, whereas the LiDAR can produce point clouds at only 20\unit{\hertz}.
The image sensor provides a 16-bit, 3840x2160 Bayer RGGB image, which is debayered, scaled to 1920x1080, and compressed for logging during autonomous operation.

\subsection{Annotation}

\begin{figure*}[t!]
    \vspace{0.2cm}
    \centering
    \includegraphics[width=\linewidth]{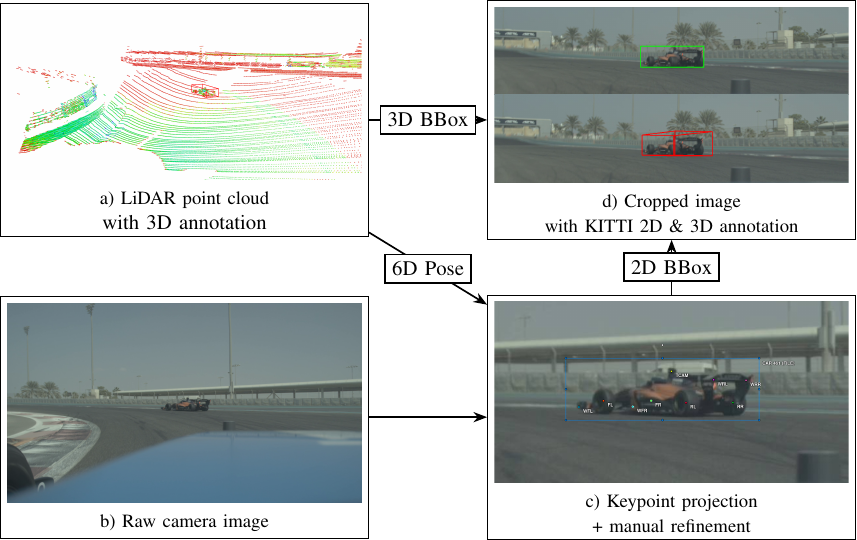}     \caption{Annotations process for new data. a) LiDAR point clouds are annotated using localization data and manual refinement. b) The matching camera image is extracted from the log. c) The annotated pose from the LiDAR data is transformed and projected into the camera frame. d) Transformed 3D (red) and 2D (green) bounding boxes are combined in the KITTI label format.}
    \label{fig:annotation}
\end{figure*}

Figure \ref{fig:annotation} shows the labeling process.
It requires manual refinement for both LiDAR point clouds and camera images.
The final annotations are converted to the Ultralytics YOLO Pose dataset format for simple model training and deployment using Ultralytics models \cite{ultralytics_ultralytics_nodate}.
The dataset consists of 976 training and 245 test frames with manually annotated  2D \& 3D bounding boxes, as well as 2D keypoints.
Frames were randomly assigned randomly to either the training or test set.
All keypoint annotations were based on projections of the annotated 3D pose from the LiDAR data.
The pose was transformed into the camera frame, expanded into 3D vehicle keypoints in the scene, and finally projected onto the image.
Afterwards, the image keypoints were refined to match image features.
Frames were annotated at 10 and 20\unit{\hertz}, with the LiDAR settings changed between sessions.
We have included labels with a maximum distance of up to 120\unit{\meter} in the dataset, which is a necessary target distance for object detection at speeds up to 260\unit{\kilo\meter\per\hour}.
On average, annotations are 58\unit{\meter} from the ego-vehicle, making the dataset challenging for depth estimation.

Images are saved in 1920x1080 resolution and rectified using the calibrated intrinsic parameters.
While the PnP solver in OpenCV accounts for distortion parameters, it is unclear whether the monocular detection algorithms trained on KITTI do so as well.
Therefore, we rectify images for both to keep the baseline comparison fair.

\subsection{KITTI Dataset Format Conversion}

To easily train benchmark models on the same data, we convert the collected data and annotations into the KITTI format.
To achieve consistent annotations, we:

\begin{enumerate}
    \item Center crop images to KITTI image size
    \item Rescale intrinsic calibration parameters
    \item Transform 3D ground-truth annotations into the  camera coordinate system
    \item Auto-generate 2D bounding box parameters from keypoint annotations
\end{enumerate}

The scenes in the dataset were chosen to avoid cropping out targets during processing.
Training, test, and validation splits are kept consistent across dataset formats to ensure fair results. 
\section{Results}

All networks were trained and evaluated on the same machine using an AMD Epyc 7313P 16-core processor, as found in the Dallara EAV24 racecar in A2RL, and a NVIDIA RTX A6000, a slightly less powerful GPU compared to the NVIDIA RTX 6000 Ada installed in the racecar.
This provides a reasonable estimate of the expected runtime of all evaluated algorithms on the target hardware.
Latency is not the only important metric for deployment; network size matters, as GPU compute must be shared with other algorithms, so lower utilization benefits the rest of the autonomous driving stack.
All YOLO networks were deployed and optimized to FP16 precision using built-in TensorRT support.

\subsubsection{MonoDETR \cite{zhang_monodetr_2023}}
MonoDETR was deployed with TensorRT to reduce latency.
This is not discussed in the original work, but it is simple to integrate into the training pipeline.
Additionally, depth partitioning and maximum range were modified to enable detection at ranges above 60m.
These changes will be referred to as MonoDETR (tuned) in the following section.
\subsubsection{MonoDTR \cite{huang_monodtr_2022}}
MonoDTR was trained using the precomputed depth ground truth from the LiDAR point cloud.
While training was successful, deployment and optimization with TensorRT were not possible with the open-source implementation because some layers are not supported by ONNX.
MonoDTR latencies are most likely similar to MonoDETR when deployed, as shown on the KITTI360 leaderboard \cite{kitti-360_kitti-360_nodate}.
\subsubsection{MonoFlex \cite{zhang_objects_2021} \& RTM3D \cite{li_rtm3d_2020,li_monocular_2021}}
MonoFlex and RTM3D/KM3D were trained on our custom, KITTI-format dataset.
During training, both models' loss exploded and diverged after a few epochs.
No meaningful detections could be produced from the trained models.

While the number of comparisons is limited, the chosen networks are current state-of-the-art leaders in benchmarks \cite{kitti-360_kitti-360_nodate,zhang_monodetr_2023}, therefore creating a challenging benchmark to overcome in both latency and accuracy. 

\subsection{2D Pose detection}

\begin{figure}[t!]
    \centering
    \includegraphics[width=\linewidth]{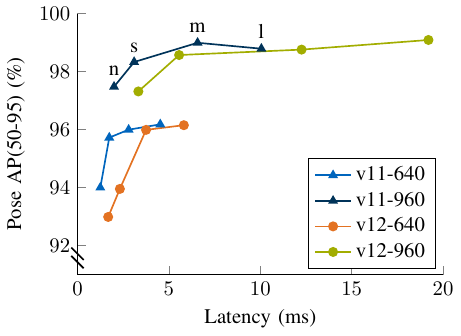}     \caption{2D detection accuracy for different YOLO model versions and sizes. For each YOLO model, we evaluate every model size from Nano (n) to Large (l).}
    \label{fig:perf2d_line_plot}
\end{figure}

Detecting the racecar's keypoints is a critical task in the first stage of the algorithm.
The Ultralytics framework was used for training the YOLO-Pose model.
All models were trained for 100 epochs, and the best epoch was selected based on the validation set.
Figure \ref{fig:perf2d_line_plot} shows the 2D accuracy of different model sizes and input image sizes.
Larger input image size matters more than model size, since many target objects are small and are compressed too much by downsampling to the input size.

\subsection{3D Detection}

\begin{figure*}[t!]
    
    \centering
    \includegraphics[width=\linewidth]{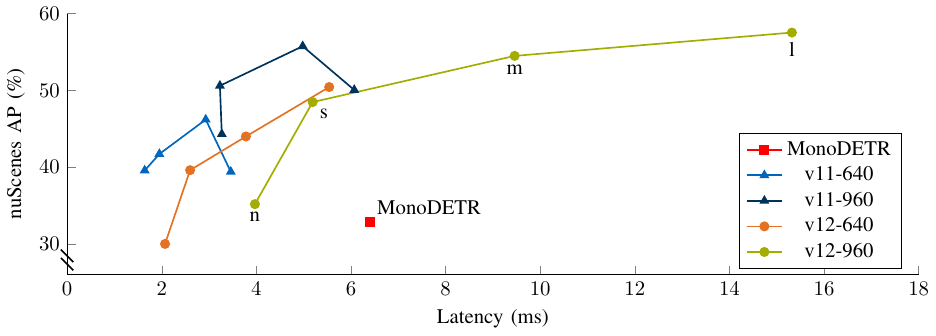}     \caption{3D detection accuracy for different YOLO model versions and sizes as well as MonoDETR. For each YOLO model, we evaluate every model size from Nano (n) to Large (l).}
    \label{fig:perf_line_plot}
\end{figure*}

\begin{table*}
    \vspace{0.2cm}
    \centering
    \caption{3D detection results on our dataset. KITTI AP refers to the BEV APR40 as evaluated by the KITTI benchmark suite. nuScenes mAP refers to the overall AP evaluated by the nuScenes devkit along with the ATE and AOE. The best result using the camera is \textbf{bold}.}
    \begin{tabular}{lcccccc}
         \textbf{Algorithm} & \textbf{Modality} & \textbf{KITTI AP\%}$\uparrow$ & \textbf{nuScenes AP\%}$\uparrow$ & \textbf{ATE}$\downarrow$ & \textbf{AOE}$\downarrow$ & \textbf{Latency (ms)}$\downarrow$ \\
CenterPoint \cite{yin_center-based_2021}
                    & L     & 51.3     &  98.4 & 0.122 & 0.219     & 28.6 \\
         VoxelNeXt \cite{chen_voxelnext_2023}
                    & L     & 94.1     & 98.4     & 0.127 & 0.075     & 31.7 \\
\hline
         \hline
         MonoDTR \cite{huang_monodtr_2022} & C & 2.8 & 12.2 & 0.839 & 0.032 & 26.2 \\
         MonoDETR \cite{zhang_monodetr_2023}
                    & C   & 2.1        & 20.4     & 0.929 & 0.043   & 6.4 \\
         MonoDETR (tuned) \cite{zhang_monodetr_2023}
                    & C   & 4.1        & 32.9     & 0.815 & 0.021   & 6.4 \\
\hline
         v11m-960 (Ours) & C & \textbf{27.6} & \textbf{55.8} & 0.664 & 0.012 & \textbf{4.9} \\
         v11m-960 (Ours, no \textit{vis} filter) & C & 24.6 & 55.5 & \textbf{0.660} & \textbf{0.011} & \textbf{4.9} \\
         v11m-960 (Ours, RANSAC) & C & 19.9 & 53.9 & 0.695 & \textbf{0.011} & 5.0 \\
         v11m-960 (Ours, Keypoint Subset) & C & 26.8 & 55.4 & 0.665 & 0.012 & \textbf{4.9} \\
         \hline
         Ground-Truth Keypoints & C & 67.7 & 85.4 & 0.159 & 0.001 & -/- \\
    \end{tabular}
    
    \label{tab:3d_results}
\end{table*}

The 3D detection evaluation compares benchmark scores and detection frequency across LiDAR and camera modalities.
Both nuScenes and KITTI AP scores are evaluated only for the single class, rather than computing an mAP.
The KITTI AP used for this evaluation is the bird's-eye-view AP with a 0.7 Intersection-over-Union threshold.
As in current implementations of object tracking for autonomous racing, the opponent's vertical position is not tracked \cite{karle_multi-modal_2023,raji_erautopilot_2024}.

LiDAR algorithms were trained on a 140\unit{\meter}x80\unit{\meter}x20\unit{\meter} voxel grid with a voxel size of 0.2\unit{\meter}x0.2\unit{\meter}x20\unit{\meter}, effectively creating pillars instead of voxels for faster inference.
This voxel grid enables the detection of distant objects in the dataset and also models the LiDAR's rough maximum range.
This parameter is a significant factor in detection frequency, so it was kept constant across all LiDAR-based algorithms.
The latency of CenterPoint and VoxelNeXt is very similar, as the OpenPCDet implementation of CenterPoint already uses sparse rather than dense convolutions, which helps reduce latency for large voxel grids.

Table \ref{tab:3d_results} presents the 3D object detection results of our approach against LiDAR algorithms and other single-camera algorithms.
MonoDTR and MonoDETR AP scores are significantly lower, as they fail to confidently predict accurate poses, especially at longer ranges.
We chose the YOLOv11m960 model to ablate, as it shows high accuracy with competitive latency.
Filtering keypoints based on their predicted visibility yields the highest accuracy, as mislabeled keypoints are discarded before the PnP solver.
However, using these occluded or invisible keypoints does not significantly reduce accuracy.
Using RANSAC to filter outliers instead of the $vis$ property did not improve accuracy while slightly increasing latency, because defining a reprojection threshold is difficult given the varying object sizes in the scene.
Lastly, we evaluated detection performance using a subset of 5 keypoints, with only the wheels and the Top-Camera used for pose estimation.
Detection performance does not drop significantly, as there are only a few occlusions where fewer than 4 keypoints are visible in the dataset.
We estimate that the full set of keypoints will matter more as occlusions in the dataset increase, given closer interaction with more opponents in the future.

It's clear that using larger network sizes and larger input images improves detection accuracy, similar to performance in 2D benchmarks.
Our method significantly outperforms MonoDETR, both in the default and tuned configurations, achieving better detection performance at range with higher confidence.
Figure \ref{fig:perf_line_plot} shows that we also achieve higher accuracy at lower latency than MonoDETR in most cases, with only the largest models exhibiting higher latency while achieving even better accuracy.
The latency of 3D pose estimation and Non-Maximum-Suppression is $\sim$0.1\unit{\milli\second}, adding minimal delay to the 2D pose model.

Additionally, we evaluate the 3D detection results of the ground-truth keypoint annotations in the last row of Table \ref{tab:3d_results}.
The results show that very accurate detection is possible; however, there are still either minor inaccuracies in the ground-truth labels or an offset between the LiDAR and camera ground truth.

While AP scores are lower than those of LiDAR algorithms, our approach exhibits significantly lower orientation errors.
This is because depth estimation is difficult and leads to relatively large translation errors; however, the estimated azimuth, elevation, and orientation are extremely precise.

\subsection{Latency}

All latency results mentioned in the previous section represent pure inference latency.
For LiDAR methods, the raw point cloud is the algorithm's input, and the copy to GPU, preprocessing, and voxelization latencies are included in the final latency measurement.
For camera-based methods, the algorithm's input is an RGB image at the correct size. For MonoDTR and MonoDETR images, the resolution was resized to the default KITTI resolution, whereas for the YOLO-based approaches, it was resized to 960x960 or 640x640.

The Debayering used for data capture is a custom, GPU-accelerated implementation that also performs brightness scaling and auto-white balance, with an average latency of $\sim$2\unit{\milli\second}.
Image scaling is handled using IsaacROS's GPU-accelerated image processing pipeline, with optional zero-copy data transfer on GPU memory via NITROS.

A disadvantage of MonoDTR and MonoDETR in this setup is the need to rectify the image in addition. SPARK does not require rectifying the input image, as the PnP-Solver can solve the Pose given calibration parameters afterward.

\section{Conclusion}

We present SPARK, a single-camera approach for 3D object detection that achieves low latency and high detection accuracy compared to state-of-the-art monocular camera detection algorithms on real-world autonomous racing data.
It enables single-camera 3D object detection at up to 200\unit{\hertz} with competitive accuracy, even at long ranges and supports object-tracking in autonomous racing with a separate and redundant detection pipeline.
Our method allows leveraging the high frame rates of the cameras installed in the autonomous racecar to produce detections with minimal delay, enabling better object tracking at racing speeds.
The evaluation shows that there is still room for improvement in 2D keypoint detection accuracy, which might be achieved with better models and more data.

\subsection*{Limitations}

A limitation of the detection method is its reliance on LiDAR ground-truth annotations for training and evaluation.
As the sensors could not be triggered exactly together, there is a residual, unavoidable offset between LiDAR and camera data in the same frame, exacerbated by the high velocities of autonomous racing.
WHile velocities are high, high angular rates cause the largest offset and are hard to compensate as they are often delayed and inaccurate in the available state estimation.
This is one of the reasons for lower detection scores than LiDAR methods; however, this will not affect tracking performance during autonomous operation, as delays can be compensated for there.

Compared to LiDAR approaches, translation errors are also high, which can be mitigated by utilizing the covariance parameters of EKF-based object tracking. While positions may exhibit some inaccuracy, orientation measurements are significantly more stable than LiDAR and can therefore be assigned lower covariances.

\section*{\uppercase{Acknowledgment}}
Dominic Ebner, as the first author, developed and implemented the approaches presented in this paper.
Markus Lienkamp made essential contributions to the conception of the research projects and revised the paper critically for important intellectual content.
Portions of the open-source code were created using the Generative AI tool ChatGPT 5 (OpenAI). The authors retain full responsibility for the accuracy, integrity, and correctness of the work.

All data was collected during season 2 of the Abu Dhabi Autonomous Racing League, organized by Aspire and the ATRC Abu Dhabi.


\end{document}